\documentclass{article}

\usepackage[nonatbib, preprint]{neurips_2019}

\usepackage{microtype}
\usepackage{graphicx}
\usepackage{subfigure}
\usepackage{booktabs} % for professional tables
\usepackage{algorithm}
\usepackage{algorithmic}
\usepackage{hyperref}

\usepackage{fncylab,enumerate,wrapfig,placeins}
\usepackage{graphicx}
\usepackage{amsmath,amsthm,amsfonts,amssymb}  
\usepackage{cuted}
\usepackage{cancel}
\usepackage{xargs}                  
\usepackage[pdftex,dvipsnames]{xcolor}
\usepackage[colorinlistoftodos,prependcaption,textsize=tiny,textwidth=2cm]{todonotes}
\newcommandx{\jianshu}[2][1=]{\todo[linecolor=red,backgroundcolor=red!25,bordercolor=red,#1]{#2}}
\usepackage{booktabs}
\usepackage{multirow}
\usepackage{xr}
\usepackage{cancel}

\makeatletter
\def\BState{\State\hskip-\ALG@thistlm}
\makeatother

\usepackage{amsmath} % assumes amsmath package installed
\def\nn{\nonumber}

\def\mc{\mathcal}

\allowdisplaybreaks

\title{Teaching Pretrained Models with Commonsense Reasoning: A Preliminary KB-Based Approach}

\author{Shiyang Li\thanks{Department of Computer Science,
University of California, Santa Barbara, CA 93106. 
Email: \texttt{shiyangli@ucsb.edu}. The work was done during an internship at Tencent AI Lab, Bellevue, WA.} \hspace{0.1in} \hspace{0.1in} Jianshu Chen\thanks{
Tencent AI Lab, Bellevue, WA, USA. Email: \texttt{jianshuchen@tencent.com} and \texttt{yudian@tencent.com}.}
\hspace{0.1in} and \hspace{0.1in} Dian Yu$^\dagger$
}

\begin{document}

\maketitle

\begin{abstract}
  Recently, pretrained language models (e.g., BERT) have achieved great success on many downstream natural language understanding tasks and exhibit a certain level of commonsense reasoning ability. However, their performance on commonsense tasks is still far from that of humans. As a preliminary attempt, we propose a simple yet effective method to teach pretrained models with commonsense reasoning by leveraging the \emph{structured knowledge} in ConceptNet, the largest commonsense knowledge base (KB). Specifically, the structured knowledge in KB allows us to construct various logical forms, and then generate multiple-choice questions requiring commonsense logical reasoning. Experimental results demonstrate that, when refined on these training examples, the pretrained models consistently improve their performance on tasks that require commonsense reasoning, especially in the few-shot learning setting. Besides, we also perform analysis to understand which logical relations are more relevant to commonsense reasoning.
\end{abstract}

\section{Introduction}

Recently, pretrained language models~\cite{devlin2018bert,radfordimproving,yang2019xlnet} have achieved great successes on various natural language understanding tasks, and they are also believed to master a certain level of commonsense reasoning abilities~\cite{liuwell,ostermann2018semeval,trinh2018simple}. Equipping machines with commonsense reasoning ability has been seen as one of the key milestones of artificial general intelligence \cite{davis2015commonsense}. However, the commonsense reasoning ability of these state-of-the-art pretrained models is still far away from that of humans~\cite{lin2019kagnet,talmor2018commonsenseqa}. One probable reason is that these models are learned from massive amounts of \emph{unstructured} texts with various language model (LM) objectives (e.g., masked language model~\cite{devlin2018bert}). That is, the commonsense reasoning capability is never explicitly taught to the pretrained models, but is implicitly acquired through modeling input texts via LM objectives. In this paper, we focus on how to \emph{explicitly} teach the pretrained models the commonsense reasoning ability.

There are several challenges in explicitly injecting commonsense reasoning capability into pretrained models. First, it is generally hard to exploit direct supervision signals for commonsense reasoning from unstructured texts, and it is also expensive, if ever possible, to create a large amount of human-labeled data for learning the commonsense reasoning ability. Second, the pretrained models do not have explicit symbolic reasoning operations; instead, the reasoning is performed implicitly through the neural network operations such as self-attention, and any knowledge relevant to reasoning is stored in the network weights. Note that the weights are only learned to fit certain input-output pairs, where the inputs to the model are natural language sentences, and the outputs are certain items to predict (e.g., masked tokens, next sentence indicator). That is, any reasoning ability has to be acquired implicitly by processing unstructured input texts during pretraining, and it is more difficult to directly supervise the reasoning path for a pretrained model.

To address these challenges, we propose a simple yet effective method to teach pretrained models with explicit commonsense reasoning abilities. The key idea is to exploit the \emph{structured} knowledge in commonsense knowledge bases (e.g., ConceptNet~\cite{speer2017conceptnet}) to generate multiple-choice questions that require commonsense reasoning. Specifically, we sample subgraphs from KB to generate various logical forms and then use text templates to generate natural language questions and candidate answers. As a result, we automatically generate a large-scale multiple-choice question answering dataset with $167$ million questions that ask about specific logical relations between different entities/concepts. These questions will be used as the additional training data to further \emph{refine} the pretrained models, which force them to learn the commonsense reasoning ability in order to answer correctly. These training inputs are already in the natural language form, which is consistent with the input of pretrained models. Therefore, it allows the model to \emph{continually} adjust its pretrained weights so that it can master more commonsense reasoning abilities; it naturally combines the power of pretrained weights from unstructured texts and the new information from structured knowledge in KB. Our experimental results show that the proposed approach consistently outperforms the baselines on commonsense reasoning tasks, especially in few-shot learning settings. In addition, we examine which logical relations are more ``commonsense'' and find that only a few simple ones are most relevant. This work is as a preliminary attempt to integrate structured commonsense knowledge into pretrained models with promising results. As we shall see, the \emph{structured} knowledge in KB allows us to \emph{systematically} construct the logical relations that we want to teach the models. We hope that our work could inspire more research towards combining structured knowledge and pretrained models.

\section{The Proposed Approach}

\begin{figure}
    \centering
    \includegraphics[width=\textwidth]{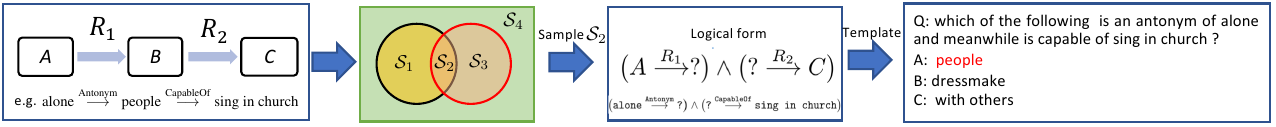}
    \caption{The generation of logical forms and multiple-choice questions in our proposed approach. The yellow and the red circles in the Venn diagram represent the sets $\mc{R}_1$ and $\mc{R}_2$, respectively.}
    \label{fig:overview_approach}
\end{figure}

The key idea of our method is to generate multiple-choice questions from different subgraphs in KB, and then we use the generated data to further \emph{refine} the pretrained models. The overall idea of the data generation process is shown in Figure \ref{fig:overview_approach}, which consists of (i) generating different logical forms from a sampled subgraph in KB, (ii) generating multiple-choice questions in natural language form. 

\subsection{Generating multiple-choice questions as the refinement data}

\paragraph{Generating logical forms}
We first sample a subgraph from KB that is in the following form:
    \begin{align}
        (A\overset{R_1}{\longrightarrow} B \overset{R_2}{\longrightarrow} C)
        \label{e:triple_pair}
    \end{align}
where $A$, $B$, and $C$ are three different entities in the KB, and $R_1$ and $R_2$ represent two different relations in the KB. For each of the above subgraph, we will construct a multiple-choice question regarding the entity $B$ in the following manner. First, introduce the following two sets: $\mc{R}_1    \!=\!  \big\{ X \in \Omega: \; A \!\overset{R_1}{\longrightarrow}\! X \big\}, \;
        \mc{R}_2    \!=\!  \big\{ X \in \Omega: \; X \!\overset{R_2}{\longrightarrow} \! C \big\}$,
where $\Omega$ denotes the entire entity set. Note that the set $\mc{R}_1$ represents the set of all (tail) entities that have relation $R_1$ with $A$, and $\mc{R}_2$ represents the set of all (head) entities that have relation $R_2$ with entity $C$. We use the two circles in the Venn diagram (Fig. \ref{fig:overview_approach}) to represent these two sets, respectively. Note from Fig. \ref{fig:overview_approach} that the entire space could be partitioned into four subsets, denoted as: $\mc{S}_1    =  \mc{R}_1 \cap \mc{R}_2^c, \;
        \mc{S}_2    =  \mc{R}_1 \cap \mc{R}_2,\;
        \mc{S}_3    =  \mc{R}_1^c \cap \mc{R}_2, \;
        \mc{S}_4    =  \mc{R}_1^c \cap \mc{R}_2^c$.
Each subset represents a certain logical relation. For example, the subset $\mc{S}_2 = \mc{R}_1 \cap \mc{R}_2$ means all the entities that have relation $R_1$ with $A$ \emph{and} have relation $R_2$ with $C$. Using these four subsets, we could compose questions that ask about all different logical relations from the subgraph in \eqref{e:triple_pair}. To see this, note that we could compose a set by either choosing or not choosing each subset $\mc{S}_i$, which leads to a total of $2^4=16$ subsets. Among them, two trivial cases are excluded: the all-chosen case (full set) and the all-not-chosen set (empty set). Therefore, there are a total of $14$ different logical relations about \eqref{e:triple_pair} that we could ask (see Appendix \ref{a:all_logical} for all the 14 logical forms). To have a more concrete example, consider the composed subset $\mc{S}_1 \cup \mc{S}_3$, then we are examining the logical relation:
    \begin{align}
        \Big(
            \big(A \overset{R_1}{\longrightarrow} ?\big) 
            \vee
            \big(? \overset{R_2}{\longrightarrow} C \big)
        \Big)
        \wedge
        \neg
        \Big(
            \big(A \overset{R_1}{\longrightarrow} ?\big) 
            \wedge
            \big(? \overset{R_2}{\longrightarrow} C \big)
        \Big)
        \label{e:logical_relation_example}
\end{align}
where $\wedge$ and $\vee$ denotes logical \texttt{AND} and logical \texttt{OR}, respectively, and $\neg$ denotes logical negation (\texttt{NOT}). This approach allows us to \emph{systematically} generate all different types of logical relations pertaining to each sampled subgraph from the KB, which even covers questions about a single relation. For example, the logical form corresponding to $\mc{S}_1 \cup \mc{S}_2$ is ``$A \overset{R_1}{\longrightarrow} ?$'', and the logical form corresponding to $\mc{S}_2 \cup \mc{S}_3$ is given by ``$? \overset{R_2}{\longrightarrow} C$'', which ask the tail entity and head entity, respectively.

\vspace{-0.5em}
\paragraph{Generating multiple-choice questions}
Now that once we have a logical form in the form of \eqref{e:logical_relation_example}, we can generate natural language questions that ask about this particular logical relation. We achieve this by using text templates. Specifically, we first create two different types of mapping, namely, \emph{affirmative mapping} and \emph{negative mapping}. The affirmative mapping is used to generate sentences with affirmative questions, while the negative mapping is used for generating negative ones. Consider the following specific example of a logical form (also shown in Figure \ref{fig:overview_approach}):
    \begin{align}
        \big(
            \texttt{alone} \overset{\texttt{Antonym}}{\longrightarrow} \texttt{?}
        \big)
        \wedge
        \big(
            \texttt{?} \overset{\texttt{CapableOf}}{\longrightarrow} \texttt{sing in church}
        \big)
        \nn
    \end{align}
where the correct answer for the missing entity is \texttt{people}. In the above logical form, the relation \texttt{CapableOf} will be mapped into ``is capable of'' using affirmative mapping. On the other hand, when there is a negation $\neg$ before the relation \texttt{CapableOf}, it will be mapped into ``is not capable of'' using a negative mapping. These obtained strings from relations will be put together with the head entities and the tail entities to generate sentences as natural as possible by using a set of simple heuristic rules. For example, the above logical relation will be mapped into the following natural language sentence: ``\emph{which of the following is an antonym of alone and meanwhile is capable of sing in church?}'' In Appendix \ref{a:all_logical}, we give examples of the possibly generated questions for all the $14$ logical relations.

\vspace{-0.5em}
\paragraph{Generating candidate answers}
The correct answer is obtained from the particular logical form that we want to examine. For example, if we want to generate a question regarding the logical form \eqref{e:logical_relation_example}, the set of correct answer is given by $\mc{S}_1 \cup \mc{S}_3$. On the other hand, for the wrong candidate answers, we will examine three different sampling strategies. The first approach is to \textit{random sample} from the all the other entities in KB~\cite{sun2019probing}. The second one is the \textit{nearest sampling}, which chooses the entity from $\{X \in \Omega: \; A \overset{R_1}{\longrightarrow} X, \forall R \neq R_1 \}$. The third sampling method is \textit{uniform sampling}: it firstly chooses wrong subset uniformly from $\mc{S}_1,\ldots,\mc{S}_4$ and then samples an entity from the selected subset.

\vspace{-0.5em}
\subsection{Refinement: teaching the pretrained models with commonsense reasoning}
\vspace{-0.5em}
To teach the pretrained models with commonsense reasoning, we further train the pretrained models on the generated multiple-choice questions to predict the correct answer, which becomes a multi-class classification problem. Afterwards, the model is finetuned on different downstream tasks. We name this step as \emph{refinement} to distinguish it from the \emph{pretraining} and the \emph{finetuning} stages.

\vspace{-1em}
\section{Experiments}
\vspace{-1em}
In this section, we examine the performance of the proposed method on different tasks and perform analysis on which logical relations are more ``commonsense''. First, we briefly describe the experimental setting, and more details could be found in Appendix \ref{a:experiment_details}. We first preprocess ConceptNet and keep 3,098,816 English-only triples. Then, we perform search on these triples and obtain a total of 167,395,947 subgraphs that are in the form of \eqref{e:triple_pair}. These subgraphs would lead to over 167 million multiple-choice questions for further refining the pretrained models. We use the \emph{uniform sampling} method to generate the wrong candidate answers unless otherwise stated. To evaluate the performance, we finetune the refined models on three downstream tasks that require strong commonsense reasoning: \textit{CommonsenseQA}~\cite{talmor2018commonsenseqa}, \textit{CosmosQA}~\cite{huang2019cosmos}, and \textit{DREAM}~\cite{sundream2018} (see Appendix \ref{a:experiment_details} for the descriptions).

\vspace{-0.5em}
\paragraph{Few-Shot Learning Performance}
In Table \ref{tab:few_shot}, we show the few-shot learning performance of our proposed method on CommonsenseQA. We consider three different types of pretrained models: BERT, GPT, and XLNet. We refine these models on our generated multiple-choice questions and then finetune them on CommonsenseQA. We compare the results to the corresponding models without the refinement process (i.e., directly finetuning on CommonsenseQA). Our method has significantly better few-shot learning performance with as large as $18\%$ absolute improvement, meaning that the refinement process effectively teaches a pretrained model commonsense reasoning even with a few finetuning samples. With full finetuning data, our method also achieves $2\%$ gain. The above results are obtained using the base models of BERT/XLNet. Additional experimental results in Appendix \ref{a:additional_experiments} show that the same performance gain could carry over to their corresponding large models.

\begin{table}[h]
\scriptsize
  \caption{Results with different size of fine-tuning data in accuracy (\%) on the CommonsenseQA development set. Data size percentages are listed in the parentheses. All results are averaged over five independent runs, with standard deviations listed inside the parentheses.}
  \label{tab:few_shot}
  \centering
  \begin{tabular}{c|cccccc|c}
  \toprule
    Data size     & 100 (1.0\%) & 200 (2.1\%)  & 400 (4.1\%)  & 800 (8.2\%) & 1600 (16.4\%) & 3200 (32.9\%) & 9741 (100\%) \\
    \midrule
    BERT~\cite{devlin2018bert} &  34.94(1.97) & 38.41(2.16) & 41.73(2.02) &  45.44(1.16) & 47.53(1.15) & 51.84(0.62) & 57.33(1.03)
    \\
    BERT + refine  & 42.54(1.27) & {\bf 44.93}(1.69) & {\bf 47.03}(0.27) & {\bf 50.58}(0.64) & {\bf 53.43}(0.85) & {\bf 54.86}(0.75) & 59.28 (0.43)
    \\
     \midrule
    GPT~\cite{radfordimproving} &  27.90(1.30) & 28.34(1.39) & 30.20(1.99) & 33.96(2.53) & 38.54(1.55) & 45.45(0.65) & 50.75(1.08)
    \\
    GPT + refine  & 37.69(0.27) & 38.56(0.37) & 40.49(0.50) & 42.49(0.44) & 44.24(0.58) & 46.50(0.35) & 51.52(0.62)
    \\
     \midrule
    XLNet~\cite{yang2019xlnet} &  25.04(0.67) & 27.44(1.08) & 29.80(2.17) & 34.27(0.89) & 38.67(1.30) & 47.14(1.25) & 57.25(1.14)
    \\
    XLNet + refine  & {\bf 43.60}(0.15) & 43.67(0.24) & 45.81(0.23) & 47.24(0.47) & 50.60(0.53) & 53.41(0.32) & {\bf 59.31}(0.44)
    \\
    \bottomrule
  \end{tabular}
\end{table}

\paragraph{Candidate answer sampling strategies} In Table \ref{tab:negative_sampling}, we show the results of different strategies to sample the wrong candidate answers on different datasets. We find that our method is relatively insensitive to different sampling strategies, and the performance varies slightly over different datasets.

\begin{table}[h]
  \caption{Performance of different strategies to sample wrong candidate answers.}
  \label{tab:negative_sampling}
  \centering
  \scriptsize
  \begin{tabular}{c|c|cc|c|c}
  \toprule
    Dataset &  Candidate answer selection   & DREAM-{dev}     & DREAM-{test} & CommonsenseQA-{dev}  & CosmosQA-{dev} \\
    \midrule
    BERT & \cancel{\phantom{nothing}} &  62.06(0.75) & 61.98(0.79) & 57.33(1.03)  & 58.34(0.82)
    \\
    BERT + refine & random sampliing &  {\bf 63.49} (0.35) & {\bf 62.89}(0.36) & 58.51(0.87)  & 58.66(0.26)
    \\
    BERT + refine & nearest sampling & 63.02(0.23) & 62.56(0.94) & {\bf 58.97}(0.76) & {\bf 59.14}(0.89)
    \\
    \bottomrule
  \end{tabular}
\end{table}

\vspace{-0.5em}
\paragraph{Which logical relations are more ``commonsense''?} To partially answer this question, we refine the pretrained BERT-base model on different subsets of logical relations from all the $14$ logical froms in Appendix \ref{a:all_logical} and report the results in Table \ref{tab:relevance_logical_commonsense}. We observe that relatively simple logical relations (\#1, \#2, \#5) (i.e., simple logical \texttt{AND} and single relation reasoning) are more relevant to commonsense; refining on just three of them achieves almost full performance (i.e., BERT + refine (all)). On the other hand, the logical forms (\#4, \#7, \#9), which require more logical compositions and negations, are less commonsense; refining on them does not improve much over the baseline BERT model. This is consistent with our intuition that commonsense should be something relatively straightforward.

\begin{table}[h]
  \caption{The relevance of different logical relations to commonsense. ``BERT + refine (1,2,5)'' means the pretrained BERT model is refined on the logical forms \#1, \#2, and \#5 defined in Appendix \ref{a:all_logical}.}
  \label{tab:relevance_logical_commonsense}
  \centering
  \scriptsize
  \begin{tabular}{c|c|ccc|c}
  \toprule
    Method    & BERT &  BERT + refine (1,2,5)    & BERT + refine (2,4,5) & BERT + refine (4,7,9)  & BERT + refine (all) \\
    \midrule
    CommonsenseQA &57.33(1.03) & { 59.10}(0.43) & 58.18(0.41) & 56.42(0.82)  & { 59.28}(0.43)
    \\
    CosmosQA & 58.34(0.82) &  59.28(0.81) & { 59.76}(0.75) & 58.86(0.74) & 58.91(0.44)
    \\
    \bottomrule
  \end{tabular}
\end{table}

\vspace{-1em}
\section{Related Work}
\vspace{-0.5em}
Structured knowledge~\cite{bollacker2008freebase,speer2017conceptnet} has been explored for question answering and reading comprehension. Most existing methods~\cite{bi2019incorporating,laugier2019encoding,li2015answering,sachan2016science,sundream2018,wang2018yuanfudao,zhong2018improving} only exploit triples relevant to questions.~\cite{sun2019probing} fine-tune models on questions constructed by predicting the head or tail mention in a triple. Very recently,~\cite{ye2019align} propose to align triples to Wikipedia sentences to form natural language questions.

\vspace{-1em}
\section{Conclusion}
\vspace{-0.5em}
In this paper, we propose a simple yet effective method to use structured knowledge (i.e., ConceptNet) to enhance the commonsense reasoning abilities of pretrained language models. The structured knowledge in KB allows us to construct various logical forms, and then generate multiple-choice questions that require commonsense logical reasoning. Experimental results demonstrate that, when refined on these training examples, these models consistently improve their performance on three datasets that require commonsense knowledge, especially in the few-shot learning setting. In the future, we are interested in designing methods to generate more diverse natural language questions instead of relying on patterns and evaluating on other recent models like RoBERTa \cite{Liu2019RoBERTaAR}.

\newpage
\bibliography{main}
\bibliographystyle{abbrv}

\newpage

\appendix
\section*{\huge Supplementary Material}
\section{Experimental Details}
\label{a:experiment_details}
In this section, we describe more details of the experimental settings and the choice of hyper-parameters.

\subsection{Experiment details of the refinement process}

\paragraph{Handling invalid logical forms.} We find that some subgraphs \eqref{e:triple_pair} sampled from KB could not generate all the $14$ logical forms in Appendix \ref{a:all_logical}. For example, if $\mc{S}_1$ is an empty set for a specific subgraph, logical form \#0 is invalid. In our implementation, we create a specific $14$-dimension $0$/$1$-mask vector for each subgraph to indicate which logical forms are valid for sampling. 

\paragraph{Efficiency considerations.} In our implementation, we use the \texttt{torch.utils.data.Dataset} class in PyTorch to generate the training data for the refinement process on-the-fly. We observe that calculating $\mc{S}_4$ is relatively time-consuming because we have to remove all the elements in $\mc{S}_1, \mc{S}_2$, and $\mc{S}_3$ from the total set for each sampled subgraph. This can be a bottleneck for the dataloader and will finally reduce the overall GPU utilization. To address this issue, we approximate $\mc{S}_4$ by the total set in our experiments, which is an efficient and relatively accurate approximation. Note that since the number of elements in $\mc{S}_1, \mc{S}_2$, and $\mc{S}_3$ is much smaller than that in the total set, the chance of sampling an element from $\mc{S}_1, \mc{S}_2$, and $\mc{S}_3$ is extremely small. Therefore, this could be an efficient and good approximation to sampling from $\mc{S}_4$.

\paragraph{Hyper-parameters for the refinement process.} When we refine BERT, GPT, and XLNet, we only train our models for one epoch. This is because we find that training too many iterations on our generated multiple-choice question answering dataset may make the model forget the pretrained language modeling capability and eventually hurt performance. We set the maximum sequence length to be $40$ during refinement as it covers most of the input texts for all three pretrained models, and we set the optimizers and the learning rates to be the same as their default values. The learning rates are set to be $2\times 10^{-5}$, $6.25 \times 10^{-5}$, and $2 \times 10^{-5}$ for BERT, GPT, and XLNet, respectively. We do not tune their hyper-parameters (e.g., learning rate) due to limited resources. Note that for GPT, language model coefficient is set to be $0$ during refining since we argue that the texts in our template datasets may not be as natural as the ones used for pretraining.

\paragraph{Experimental setting for Table \ref{tab:relevance_logical_commonsense}.} For BERT + refine (all), we sample over all valid logical forms according to a uniform distribution. For BERT + refine (1,2,5), BERT + refine (2,4,5), and BERT + refine (4,7,9), logical forms are uniformly sampled over (\#1, \#2, \#5), (\#2, \#4, \#5), and (\#4 , \#7, \#9), respectively. The training procedures follow the same hyper-parameters described above. For the finetuning process in all experiments, we train a model five times and report their mean values and standard deviations.

\subsection{Description of the downstream tasks}
\textbf{CommonsenseQA} dataset consists of 12,247 multi-choice question answer pairs with one correct answer and four incorrect
answers requiring commonsense reasoning capability \cite{talmor2018commonsenseqa}. This dataset has two kinds of splits, namely \textit{question concept split}  and \textit{random split} \cite{talmor2018commonsenseqa}, and our experiments are
conducted on the official random split. For few-shot learning experiments, we allow models to train more epochs to make sure that they converge. Specifically, for training data size in $\{100, 200, 400, 800, 1600, 3200\}$ , we train models with $\textit{epoch} \in \{100, 50, 25, 12, 8, 8\}$, respectively and keep other settings fixed. For training on the whole dataset, we follow similar settings of officially released code \footnote{\url{https://github.com/jonathanherzig/commonsenseqa/tree/master/bert}.}. For a fair comparison between baselines and our refining methods, we keep their epochs, batch sizes, and other settings the same. The only differences are parameters where baselines utilize officially pretrained models, and ours use checkpoints during the proposed refining processes. 

\textbf{CosmosQA} dataset consists of $35.6$K multiple-choice reading comprehension problems requiring commonsense reasoning capability from given contexts \cite{huang2019cosmos}. Therefore, it is an appropriate dataset for testing commonsense reasoning of models. We finetune baselines and our proposed methods for four epochs with learning rate 2e-5 and batch size of $36$. We evaluate models on the development set in every epoch and report the best performance for each experiment.

\textbf{DREAM} dataset consists of 10,197 multiple-choice dialogue-based machine reading comprehension problems and answering $34\%$ of them needs commonsense reasoning knowledge \cite{sundream2018}. We adapt the officially released BERT-based source code on DREAM and choose the same setting as the repository.\footnote{\url{https://github.com/nlpdata/mrc_bert_baseline}.} Similar to \textit{CosmosQA}, we evaluate development set in every epoch and report the best performance and its corresponding test set accuracy for each experiment.

\section{All Logical Forms with Example Multiple-Choice Questions}
\label{a:all_logical}
In this appendix, we show all the $14$ logical relations that could be sampled from a particular triple pair, and the examples for the corresponding generated multiple-choice questions. Specifically, we consider the following example of triple pair:
    \begin{align}
        (\text{arise}\overset{\text{Antonym}}{\longrightarrow} \text{sit}, \;  \text{sit} \overset{\text{RelatedTo}}{\longrightarrow} \text{sit up})
        \nn
        % \label{e:triple_pair}
    \end{align}
Then, all the $14$ logical forms and the corresponding example questions are given below, where the correct answer is highlighted in red and bolded:
\begin{itemize}
    \item \underline{logical form \#0}: $\mc{S}_1$ 

    \begin{align}
            \big(A \overset{R_1}{\longrightarrow} ?\big) 
        \wedge
        \neg
            \big(? \overset{R_2}{\longrightarrow} C \big)
    \end{align}
Q: \textit{which of the following is an antonym of arise and meanwhile is not related to sit up ?} \\
A: {\color{red} \bf  \textit{set}} \\
B: \textit{fancifying} \\ 
C: \textit{storing space shuttle}

    \item \underline{logical form \#1}: $\mc{S}_2$ 

    \begin{align}
            \big(A \overset{R_1}{\longrightarrow} ?\big) 
        \wedge
            \big(? \overset{R_2}{\longrightarrow} C \big)
    \end{align}
    
Q: \textit{which of the following is an antonym of arise and meanwhile is related to sit up ?} \\
A: {\color{red} \bf  \textit{sit}} \\
B: \textit{sitting up} \\ 
C: \textit{stand up}

\item \underline{logical form \#2}: $\mc{S}_1 \cup \mc{S}_2$ 

    \begin{align}
            \big(A \overset{R_1}{\longrightarrow} ?\big) 
    \end{align}

Q: \textit{which of the following is an antonym of arise ?} \\
A: \textit{promegapoietin} \\ 
B: \textit{sleigher} \\
C: {\color{red} \bf  \textit{set}} \\
\item \underline{logical form \#3}:  $\mc{S}_3 $ 

    \begin{align}
            \neg\big(A \overset{R_1}{\longrightarrow} ?\big) 
        \wedge
            \big(? \overset{R_2}{\longrightarrow} C \big)
    \end{align}

Q: \textit{which of the following is not an antonym of arise and meanwhile is related to sit up ?} \\
A: \textit{craftist} \\ 
B: \textit{queer anarchism} \\
C: {\color{red} \bf  \textit{stand up}} \\

\item \underline{logical form \#4}:  $\mc{S}_1 \cup \mc{S}_3 $
   \begin{align}
        \Big(
            \big(A \overset{R_1}{\longrightarrow} ?\big) 
            \vee
            \big(? \overset{R_2}{\longrightarrow} C \big)
        \Big)
        \wedge
        \neg
        \Big(
            \big(A \overset{R_1}{\longrightarrow} ?\big) 
            \wedge
            \big(? \overset{R_2}{\longrightarrow} C \big)
        \Big)
    \end{align}

Q: \textit{which of the following is an antonym of arise or is related to sit up, but not both of them ?} \\
A: \textit{sit down} \\ 
B: \textit{make refreshing dessert} \\
C: {\color{red} \bf  \textit{lower}} \\

\item \underline{logical form \#5}: $\mc{S}_2 \cup \mc{S}_3 $     \begin{align}
            \big(? \overset{R_2}{\longrightarrow} C \big)
    \end{align}

Q: \textit{which of the following  is related to sit up ?} \\
A: \textit{lay} \\ 
B: {\color{red} \bf  \textit{sitting up}} \\
C: \textit{descend} \\

\item \underline{logical form \#6}: $\mc{S}_1 \cup \mc{S}_2  \cup \mc{S}_3 $ 

 \begin{align}
            \big(A \overset{R_1}{\longrightarrow} ?\big) 
        \vee
            \big(? \overset{R_2}{\longrightarrow} C \big)
    \end{align}

Q: \textit{which of the following is an antonym of arise or is related to sit up ?} \\
A: \textit{marksberrys} \\ 
B: {\color{red} \bf  \textit{sit down}} \\
C: \textit{previsive} \\

\item \underline{logical form \#7}: $\mc{S}_4$ 
   \begin{align}
           \neg \big(A \overset{R_1}{\longrightarrow} ?\big) 
            \wedge
            \neg \big(? \overset{R_2}{\longrightarrow} C \big)
    \end{align}
    
Q: \textit{which of the following is not an antonym of arise and is not related to sit up ?} \\
A: \textit{crunch} \\ 
B: {\color{red} \bf  \textit{millikin}} \\
C: \textit{sit} \\

\item \underline{logical form \#8}: $\mc{S}_1 \cup \mc{S}_4 $  

   \begin{align}
            \neg \big(? \overset{R_2}{\longrightarrow} C \big)
    \end{align}

Q: \textit{which of the following  is not related to sit up ?} \\
A: \textit{sit down} \\ 
B: {\color{red} \bf  \textit{simpliciter}} \\
C: \textit{crunch} \\

\item \underline{logical form \#9}: $\mc{S}_2 \cup \mc{S}_4 $ 
    \begin{align}
        \Big(
            \big(A \overset{R_1}{\longrightarrow} ?\big) 
            \wedge
            \big(? \overset{R_2}{\longrightarrow} C \big)
        \Big)
        \vee
        \Big(
            \neg \big(A \overset{R_1}{\longrightarrow} ?\big) 
            \wedge
            \neg \big(? \overset{R_2}{\longrightarrow} C \big)
        \Big)
    \end{align}

Q: \textit{which of the following is an antonym of arise and is related to sit up, or neither of them ?} 
A: \textit{fall down} \\ 
B: {\color{red} \bf  \textit{cremators}} \\
C: \textit{lower} \\

\item \underline{logical form \#10}: $\mc{S}_1 \cup \mc{S}_2  \cup \mc{S}_4 $ 

   \begin{align}
            \big(A \overset{R_1}{\longrightarrow} ?\big) 
        \vee
          \neg  \big(? \overset{R_2}{\longrightarrow} C \big)
    \end{align}

Q: \textit{which of the following is an antonym of arise or is not related to sit up ?} \\
A: \textit{sitting up} \\ 
B: {\color{red} \bf  \textit{sit down}} \\
C: \textit{stand up} \\

\item \underline{logical form \#11}: $\mc{S}_3 \cup \mc{S}_4 $ 

   \begin{align}
           \neg \big(A \overset{R_1}{\longrightarrow} ?\big) 
    \end{align}

Q: \textit{which of the following is not an antonym of arise ?} \\
A: \textit{lay down} \\ 
B: {\color{red} \bf  \textit{free criminals}} \\
C: \textit{abed} \\
\\
\item \underline{logical form \#12}: $\mc{S}_1 \cup \mc{S}_3  \cup \mc{S}_4 $ 

   \begin{align}
        \neg    \big(A \overset{R_1}{\longrightarrow} ?\big) 
        \vee
          \neg  \big(? \overset{R_2}{\longrightarrow} C \big)
    \end{align}

Q: \textit{which of the following is not an antonym of arise or is not related to sit up ?} \\
A: \textit{sit down} \\ 
B: \textit{sit down} \\
C: {\color{red} \bf  \textit{snub line}} \\

\item \underline{logical form \#13}: $\mc{S}_2 \cup \mc{S}_3  \cup \mc{S}_4 $ 

   \begin{align}
        \neg    \big(A \overset{R_1}{\longrightarrow} ?\big) 
        \vee
          \big(? \overset{R_2}{\longrightarrow} C \big)
    \end{align}

Q: \textit{which of the following is not an antonym of arise or is related to sit up ?} \\
A: \textit{lower} \\ 
B: \textit{fall} \\
C: {\color{red} \bf \textit{sit down}} \\

\end{itemize}

\section{Additional Experiments}
\label{a:additional_experiments}

In this section, we list additional experiments with the BERT-large model. All results are averaged over five independent runs, with standard deviations listed inside the parentheses.

\begin{table}[h]
  \caption{Performance of our proposed method on BERT-large vs. on BERT-base, evaluated on the DREAM dataset. The performance metric is prediction accuracy (\%). The standard deviations are listed inside the parenthesis.}
  \label{tab:model_size}
  \centering
  \begin{tabular}{cccc}
  \toprule
    Data   & Candidate answer selection & DREAM-dev     & DREAM-test \\
    \midrule
    BERT-base & \cancel{\phantom{Nothing}} & 62.06(0.75)& 61.98(0.79) 
    \\
    BERT-base + refine & random sampling &  63.49(0.35) & 62.89(0.36)
    \\
    \midrule
    BERT-large & \cancel{\phantom{Nothing}} & 65.62(0.62) & 66.07(1.00)
    \\
    BERT-large + refine & random sampling  & {\bf 67.11}(0.48) & {\bf 67.36}(0.63)
    \\
    \bottomrule
  \end{tabular}
\end{table}

\end{document}